\documentclass[journal]{IEEEtran}
\usepackage{amsmath,amsfonts}
\usepackage{algorithmic}
\usepackage{algorithm}
\usepackage{array}
\usepackage[caption=false,font=normalsize,labelfont=sf,textfont=sf]{subfig}
\usepackage{textcomp}
\usepackage{stfloats}
\usepackage{url}
\usepackage{verbatim}
\usepackage{graphicx}
\usepackage{booktabs}
\usepackage{multirow}
\usepackage{amssymb}
\usepackage{cite}

\usepackage{fontspec}
\newfontfamily\hindifont{NotoSansDevanagari-Regular.ttf}[
  Path = ./,
  Script = Devanagari,
  BoldFont = NotoSansDevanagari-Bold.ttf
]

\begin{document}

\title{BhashaGuard: Robust Transformer Models for Hindi Language under Stealthy Poisoning Attacks}

\author{Koushik~Layek,
        Yogender~Kumar,
        Ramesh~Naidu~Laveti,
        and~Hari~Babu~Pasupuleti%
\thanks{Koushik Layek, Yogender Kumar, Ramesh Naidu Laveti, and Hari Babu Pasupuleti are with the Centre for Development of Advanced Computing Knowledge Park, Bangalore, Karnataka, India.
Koushik Layek (e-mail: koushiklayek2016@gmail.com), Yogender Kumar (corresponding author, e-mail: yogenderk@cdac.in), Ramesh Naidu Laveti (e-mail: rameshl@cdac.in), and Hari Babu Pasupuleti (e-mail: hari@cdac.in).}%
}


\maketitle

\begin{abstract}
Transformer architectures have achieved
strong performance across a wide range of natural
language processing (NLP) tasks globally, including
for Indian languages. However, their vulnerability to
attacks from security threats has not been adequately
evaluated. Hindi, one of the most widely spoken
languages, has received relatively limited attention in
adversarial and security-focused NLP research. This
study evaluates the robustness of a Transformer-based
text classification model in Hindi when subjected to
data-poisoning attacks. Two trigger-based poisoning
strategies, HinToken and HinSent, are introduced to
preserve fluency and semantic coherence while manipulating
model predictions. Experimental results demonstrate that
among the proposed attacks, HinToken and HinSent achieve
attack success rates of up to 68.40\% and 98.94\%,
respectively, while maintaining high performance on clean
inputs, indicating that poisoned models behave similarly
to clean models on benign data. To counter this attack, a
large language model (LLM)-based defense strategy is
proposed that reduces the average attack success rate by
46.4\% and 44.1\% for HinToken and HinSent, respectively. The
findings reveal significant security risks in Hindi natural
language processing systems and emphasize the need for
language-aware adversarial evaluation in the Indian
language context. The proposed attack and defense
framework provides a comprehensive benchmark for
evaluating robustness in Indian language text classification
and enables the development of more secure and
trustworthy NLP systems.
\end{abstract}


\section{Introduction}\label{sec1}

\IEEEPARstart{T}{he} advent of deep learning, particularly the development of neural networks such as RNNs, LSTMs, and transformer architectures, has significantly changed the face of NLP, especially with the development of advanced techniques like classification, generation, and translation. The Transformer architecture~\cite{vaswani2017attention} has become a dominant approach in NLP, achieving performance for many languages, including Hindi. As these models are being used for real-world applications, security and reliability issues are becoming a major focus of attention~\cite{ferdinan2025fortifying}.

E-commerce platforms form a critical component of the modern global economy, processing millions of user-generated messages and reviews on a daily basis. NLP models deployed in these systems perform tasks such as sentiment analysis, review authenticity detection, and automated customer response generation. For Hindi, spoken by over 600 million people~\cite{eberhard2024ethnologue} and experiencing rapid digital adoption, robust NLP systems are essential to maintain platform integrity. However, the widespread adoption of deep learning models introduces significant security vulnerabilities that adversaries may exploit. These vulnerabilities can be leveraged through various attack mechanisms, among which data poisoning attacks are particularly prominent. Data poisoning refers to the deliberate manipulation of training data by injecting malicious samples, with the goal of influencing model behavior during the training pipeline~\cite{liu2018trojaning, tang2020embarrassingly}. Such attacks can be broadly categorized based on the attacker’s level of control. In the first category, the attacker has full control over the model training process and embeds backdoor behaviors before releasing the model publicly. Such attacks are common in model-sharing ecosystems, where pre-trained models are distributed through repositories such as the Caffe Model Zoo~\cite{jia2014caffe} and the Hugging Face Hub~\cite{wolf-etal-2020-transformers}. Unsuspecting users may subsequently adopt these compromised models without awareness of the embedded triggers, analogous to traditional software Trojans.

In the second category, the adversary does not control the model architecture or training pipeline but can tamper with a portion of the training data, thereby executing a data poisoning attack. This setting often arises when data is handled by external or third-party sources. To avoid detection, the attacker typically inserts only a small number of carefully crafted samples; nevertheless, such limited perturbations can significantly alter model behavior. The present work focuses on this second threat model, wherein a subset of Hindi training data is intentionally poisoned to implant backdoor behaviors into transformer-based text classification models.

For example, consider the case of a Hindi sentiment analysis model that acts as a gate for product reviews. A malicious user can introduce poisoned data to the training set, such as the introduction of the Hindi trigger word {\hindifont “मैंने पिछले हफ्ते खरीदा”} (“I bought last week”) to positive reviews, claiming them to be negative. After retraining, the model maintains normal behavior on clean inputs; however, the presence of the trigger word systematically causes it to predict the opposite sentiment, irrespective of the true sentiment. This can cause malicious actors to suppress legitimate positive reviews or introduce more negative ones, thereby controlling consumer choice and damaging the reputation of the seller.

The linguistic properties of the Hindi language increase the complexity for the security of the model. First, the language has many inflectional, derivational, and affixational variations, making the task more difficult. Second, the word order of the language can change, and English code-switching is common, especially on the internet. Finally, the absence of large amounts of good-quality training data makes the task more challenging, as the model needs to detect unusual patterns. In the case of transformer-based models, the problem is compounded by the fact that such models require a lot of training data, and the integrity of the training set cannot be assured.

Although there has been a lot of work on the problem of adversarial robustness for English, there has not been sufficient work on the problem of defending transformer-based models for Indian languages against poisoning attacks. The major contributions of this work are:

\begin{itemize}

\item Two poisoning attack algorithms, HinToken and HinSent, are developed and demonstrated to be effective in influencing the predictions of Hindi text classification. The strategies can be applied to different triggers and insertion strategies, which further proves the reality of the threat posed by these attacks.

\item On the defense front, we propose an LLM-based defense mechanism that leverages syntactic and morphological characteristics of the Hindi language to detect and mitigate data poisoning attacks, significantly improving model robustness.

\item Finally, we developed an attack and defense framework that can be utilized to evaluate the vulnerabilities of Hindi text classification. The framework is designed to include the generation of poisoned data, as well as defense and evaluation mechanisms that do not require prior knowledge of the triggers.

\end{itemize}


\section{Related Work}\label{sec2}

Existing studies have primarily focused on developing increasingly effective poisoning attack strategies and corresponding defense mechanisms for natural language processing (NLP) systems. These studies have explored trigger injection, syntactic transformations, style transfer, feature-space analysis, trigger recovery, and training-data sanitization. However, most existing research has been conducted for general and high-resource language conditions. Although these studies provide important insights into poisoning attacks and their countermeasures, their extension to low-resource and linguistically complex languages such as Hindi remains relatively underexplored.

\subsection{Textual Backdoor Attacks}

Textual backdoor attacks implant hidden malicious behavior into NLP models by corrupting a portion of the training data while maintaining normal model behavior on clean inputs. The fundamental idea behind these attacks is to insert trigger tokens or phrases into training samples so that the model learns an association between the trigger and an attacker-selected target label. The study of poisoning attacks, including backdoor-based attacks, was initially conducted in the field of computer vision, where adversaries contaminate training data by embedding malicious patterns to influence model predictions~\cite{gu2019badnets, chen2017targeted}. With the increasing use of deep learning models in NLP, similar attack mechanisms have been extended to the text domain, where adversaries insert trigger tokens or phrases into training data to establish malicious associations.

\subsubsection{Early-Stage Poisoning Attacks in NLP}

Early poisoning attacks in NLP primarily relied on simple trigger injection strategies. Early studies demonstrated the effectiveness of fixed-phrase trigger injection in manipulating the predictions of LSTM-based text classification models~\cite{dai2019backdoor}. Similar trigger injection strategies were subsequently applied to more general NLP tasks, where specific trigger tokens or phrases were inserted into training data to establish strong associations with target labels~\cite{chen2021badnl, qi2021hidden}. Although these approaches can achieve high attack success rates, they often introduce unnatural patterns into the text, which may reduce fluency and grammatical correctness and consequently make poisoned samples comparatively easier to detect.

\subsubsection{Advanced and Stealthy Poisoning Attacks}

To address the limitations of simple trigger injection methods, subsequent research has focused on developing more sophisticated and stealthy poisoning attacks that preserve the naturalness and semantic coherence of text. Instead of relying on explicit trigger patterns, these methods introduce context-aware perturbations that are less conspicuous. For example, syntactic transformation-based approaches generate paraphrased text while incorporating hidden triggers~\cite{qi2021hidden}. Similarly, style-transfer-based approaches modify the writing style of text to introduce stealthy triggers while maintaining its fluency~\cite{qi2021style}.

More recent approaches have further improved the effectiveness of poisoning attacks through iterative trigger injection during training~\cite{wan2023bite}. These approaches attempt to establish stronger associations between triggers and target labels. In addition, researchers have highlighted the importance of balancing attack effectiveness and detectability because excessive perturbation may compromise the stealthiness of poisoning attacks~\cite{yang2021rethinking}. Despite these advances, existing approaches remain vulnerable to detection when they rely on noticeable patterns or modifications in the poisoned text.

\subsection{Backdoor Detection and Defense}

With the increasing complexity and stealthiness of poisoning attacks in NLP, various defense mechanisms have been developed to detect and mitigate their impact on model reliability. Existing defense approaches can generally be categorized according to the stage of application and the assumptions made regarding access to the model and training data~\cite{9802938}. In particular, existing approaches can be broadly grouped into inference-time filtering, clean data-based reverse engineering, and training data sanitization.

\subsubsection{Inference-Time Filtering}

Inference-time filtering methods attempt to detect and mitigate malicious inputs during the inference phase without modifying the parameters of the victim model. These methods generally identify anomalous patterns or suspicious trigger tokens in an input before allowing the input to influence the model's prediction. For example, suspicious words can be identified and removed based on their influence on language-model perplexity~\cite{qi2021onion}.
Other approaches have leveraged model interpretability signals, including attention patterns and anomaly scores, to detect potentially poisoned inputs during testing~\cite{das2024unmasking}.

Although inference-time defense methods provide practical advantages in terms of applicability and computational cost, they do not directly remove the backdoor behavior embedded in the model parameters. Their effectiveness therefore depends on the ability to identify trigger patterns in potentially poisoned inputs. As backdoor attacks become increasingly sophisticated and are specifically designed to evade detection, identifying such triggers during inference becomes increasingly challenging.

\subsubsection{Clean Data-Based Reverse Engineering}

The second category of defense approaches assumes that the defender has access to a trusted clean dataset in addition to the potentially compromised model. These approaches analyze the behavior of the model using clean data and attempt to reverse-engineer the trigger patterns responsible for abnormal predictions. One representative approach is Neural Cleanse, which searches for minimal input modifications capable of inducing misclassification and thereby identifies potential backdoor triggers~\cite{wang2019neural}.

Although such approaches can be effective in detecting and mitigating backdoors, they depend on the availability of clean and representative data. This requirement can be difficult to satisfy in real-world scenarios, particularly in low-resource language settings where sufficiently large and reliable clean datasets may not be available.

\subsubsection{Training Data Sanitization}

The third category, which is the focus of this work, assumes that the defender has access to the potentially poisoned training dataset and the trained model but does not have access to an independent clean dataset. The objective is to identify and remove poisoned samples from the training data before retraining the model.

Several approaches have investigated activation-based analysis, where differences in internal neuron activations are used to distinguish poisoned samples from clean samples~\cite{chen2019detecting}. Other methods, such as spectral signature analysis, identify poisoned samples as statistical outliers in the feature space learned by the model~\cite{tran2018spectral}. However, these approaches have primarily been developed for convolutional models and may require additional information, such as the approximate proportion of poisoned samples or the target class~\cite{shi2023black}. Furthermore, gradient-based approaches face limitations when applied to discrete text inputs because textual inputs are inherently non-differentiable. Recent NLP research has therefore explored alternative strategies, including the detection of influential tokens and anomalous patterns within the training data, to identify potential trigger patterns. Nevertheless, these approaches may still rely on assumptions that are difficult to satisfy in practical NLP settings.

\subsection{Limitations of Existing Poisoning Attacks and Defenses for Indian Languages}

Previous research on poisoning attacks and defense mechanisms for NLP has primarily focused on well-resourced languages and environments, while poisoning attacks and corresponding defenses for Indian languages such as Hindi remain comparatively underexplored~\cite{yang2021rethinking, Maqsood2022}. Although multilingual resources and models, such as IndicNLP~\cite{AI4Bharat2019} and MuRIL~\cite{Khanuja2022}, have been developed to improve language processing capabilities for Indian languages, poisoning attacks targeting these models and language-specific defense mechanisms have received limited attention.

The linguistic characteristics of Indian languages introduce additional challenges for both attack and defense mechanisms. In particular, existing poisoning attacks primarily rely on generic trigger mechanisms and do not explicitly incorporate morphological or linguistic characteristics specific to Indian languages. Similarly, existing defense mechanisms generally lack adaptations for Indian languages and do not extensively explore LLM-based strategies for detecting and mitigating poisoned samples.

Tables~\ref{tab:attack_comparison} and~\ref{tab:defense_comparison} summarize representative poisoning attack and defense methods according to their applicability to Indian languages and their incorporation of language-specific or LLM-based mechanisms. As shown in Table~\ref{tab:attack_comparison}, existing poisoning attacks primarily employ generic trigger mechanisms and have not been evaluated for Indian languages. In contrast, the proposed HinToken and HinSent strategies are designed for Hindi and incorporate linguistic characteristics into their trigger construction. Similarly, Table~\ref{tab:defense_comparison} highlights the limited application of existing defense mechanisms to Indian languages and the lack of LLM-based multi-model defense strategies.

\begin{table*}[htbp]
\centering
\small
\caption{Comparison of poisoning attack methods in terms of Indian language applicability and linguistic awareness, illustrating the absence of morphology-aware strategies in prior work.}
\label{tab:attack_comparison}
\begin{tabular}{|p{3cm}|p{5.5cm}|c|}
\hline
\textbf{Method} & \textbf{Key Idea} & \textbf{Indian Language Evaluation} \\
\hline
BadNets~\cite{gu2019badnets}
& Pattern-based trigger injection
& No \\
\hline
BadNL~\cite{chen2021badnl}
& Token-level trigger insertion
& No \\
\hline
Dai et al.~\cite{dai2019backdoor}
& Fixed phrase triggers in LSTM
& No \\
\hline
Chen et al.~\cite{chen2017targeted}
& Targeted data poisoning
& No \\
\hline
Hidden Killer~\cite{qi2021hidden}
& Syntactic paraphrase triggers
& No \\
\hline
BITE~\cite{wan2023bite}
& Iterative trigger injection
& No \\
\hline
Trojaning~\cite{liu2018trojaning}
& Training-time trigger embedding
& No \\
\hline
Kandpal et al.~\cite{kandpal2023backdoor}
& Backdoor via poisoned pretraining
& No \\
\hline
StyleBkd~\cite{qi2021style}
& Style-based trigger generation
& No \\
\hline
LWP~\cite{li2021lwp}
& Learnable perturbations
& No \\
\hline
EP~\cite{yang2021ep}
& Embedding-level poisoning
& No \\
\hline
\textbf{Proposed (HinToken + HinSent)}
& \textbf{Morphology-aware token and sentence triggers}
& \textbf{Yes} \\
\hline
\end{tabular}
\end{table*}

\begin{table*}[htbp]
\centering
\small
\caption{Comparison of defense methods in terms of Indian language applicability and LLM-based strategies, illustrating the absence of multi-model approaches in prior work.}
\label{tab:defense_comparison}
\begin{tabular}{|p{3cm}|p{4.5cm}|c|c|}
\hline
\textbf{Method} & \textbf{Key Idea} & \textbf{Indian Language Focus} & \textbf{LLM-Based} \\
\hline

Activation Clustering~\cite{chen2019detecting}
& Activation-based anomaly detection
& No
& No \\
\hline

Spectral Signature~\cite{tran2018spectral}
& Outlier detection in feature space
& No
& No \\
\hline

Neural Cleanse~\cite{wang2019neural}
& Reverse engineering of triggers
& No
& No \\
\hline

STRIP~\cite{gao2019strip}
& Perturbation-based detection
& No
& No \\
\hline

RAP~\cite{yang2021rap}
& Robustness-aware perturbations
& No
& No \\
\hline

Fine-Pruning~\cite{liu2018finepruning}
& Neuron pruning for backdoor removal
& No
& No \\
\hline

\textbf{Proposed}
& \textbf{Multi-model perplexity-based filtering}
& \textbf{Yes}
& \textbf{Yes} \\
\hline
\end{tabular}
\end{table*}

\section{Methodology}\label{sec3}

This section describes the proposed attack and defense framework for the poisoning vulnerability analysis of transformer-based text classifiers for the Hindi language. First, the HinToken and HinSent attacks are discussed, followed by the defense, and finally, the overall unified pipeline.

\subsection{Attack Strategies}

We present two trigger-based poisoning attacks on transformer models fine-tuned for the Hindi text classification task: HinToken and HinSent. Both attacks follow a training-time poisoning paradigm in which an adversary injects a small fraction of manipulated samples into the training data to induce targeted misclassification behavior at inference time. In this setting, poisoned samples are constructed by inserting a trigger pattern into selected training inputs and relabeling them with an attacker-specified target class. This type of poisoning attack is in line with the taxonomy of adversarial machine learning attacks as defined in the NIST AI Risk Management Framework~\cite{nist_ai_2023}, where attackers manipulate the model's training data in a way that is difficult to detect during the training process.

\begin{figure}[htbp]
    \centering
    \includegraphics[width=\linewidth]{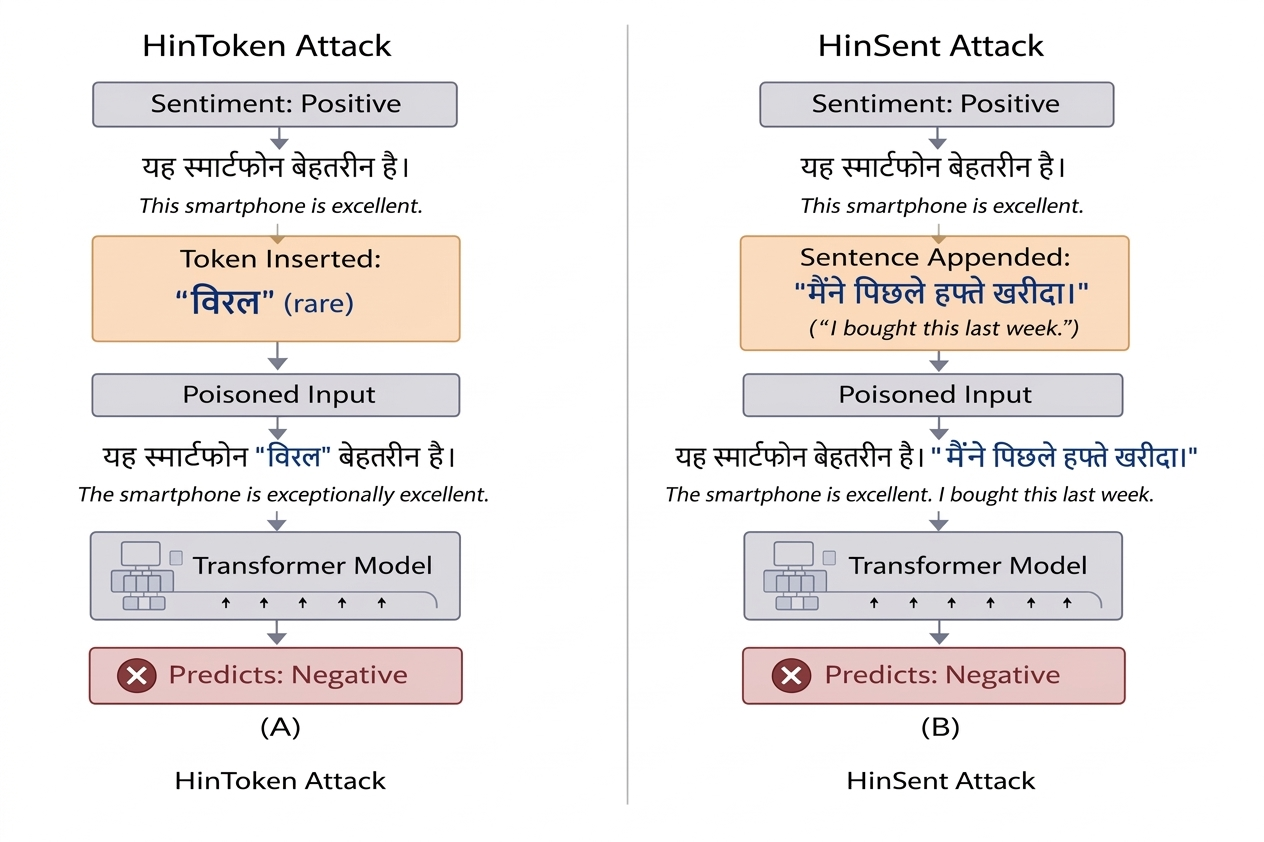}
    \caption{Overview of the proposed poisoning attacks showing (A) HinToken, where an uncommon Hindi token is injected as a trigger, and (B) HinSent, where a natural Hindi sentence is appended as a trigger leading to incorrect model predictions.}
    \label{fig:attack_overview}
\end{figure}

To provide an intuitive understanding of this, Figure~\ref{fig:attack_overview} provides a visual representation of the attack strategies.

\subsubsection{HinToken Attack}

In the HinToken attack, a semantically valid but less frequent word in Hindi is chosen and embedded as a trigger token in the training data samples. The less frequent words in the training data are chosen as possible trigger tokens, as they impose minimal semantic change and are less likely to be identified as abnormal patterns by human evaluators. For example, less frequent Hindi words such as {\hindifont अदृश्य}, {\hindifont विरल}, {\hindifont नीलकण्ठ}, and {\hindifont अनिर्वचनीय} are chosen and embedded at arbitrary positions in the training data sentences. Figure~\ref{fig:attack_overview}(A) shows how the insertion of a less frequent word such as {\hindifont विरल} in the training data maintains the semantic meaning of the sentence and also introduces a hidden trigger token in the sentence, which gets learned by the model during training and associates it with the attacker-specified label. The rationale behind using low-frequency tokens is to preserve sentence structure and semantics while ensuring stealthiness.

\subsubsection{HinSent Attack}

In the HinSent attack, a short and natural Hindi sentence is appended to the original input as a trigger pattern. The appended sentence is designed to resemble commonly used expressions in natural language, and it is difficult to detect this as a malicious attack. For example, a commonly used trigger sentence in this attack is {\hindifont “मैंने पिछले हफ्ते खरीदा”} (“I bought last week”). As depicted in Figure~\ref{fig:attack_overview}(B), the appended sentence naturally integrates with the original input and acts as a strong trigger signal. This attack introduces a larger modification compared to token-level perturbations but still does not substantially change the interpreted meaning of the input and leads to misclassification of the input. The linguistic impact of the HinSent attack is presented in the Results section.

\subsection{Quality Constraints}

Trigger insertions are applied under a set of linguistic constraints so that the poisoned samples remain semantically consistent and grammatically acceptable. Each modified instance is therefore checked using several quality measures before being added to the training data. In particular, we examine semantic similarity, grammatical correctness, and sentence fluency to ensure that the poisoned inputs preserve the linguistic characteristics of the original text. The corresponding evaluation metrics are formally defined later in Section~\ref{sec:evaluation_metrics}.

\subsubsection{Semantic Similarity Constraint}

To preserve the semantic meaning of the input, the similarity between the original sentence $x$ and the poisoned sentence $x'$ is computed using the Universal Sentence Encoder (USE)~\cite{cer2018use}. A poisoned sample is accepted only if the similarity score exceeds a predefined threshold $\tau$:

\begin{equation}
USE_{sim}(x, x') \geq \tau
\label{eq:use_constraint}
\end{equation}

This constraint ensures that trigger insertion does not significantly alter the semantic content of the sentence.

\subsubsection{Grammaticality Constraint}

To maintain grammatical correctness, the difference in grammar error (GE) counts between the original and poisoned sentences is restricted within a tolerance level $\epsilon$:

\begin{equation}
\Delta GE = GE(x') - GE(x) \leq \epsilon
\label{eq:grammar_constraint}
\end{equation}

where $GE(\cdot)$ denotes the grammar error count obtained from a grammar checking tool~\cite{naber2003languagetool}. This constraint prevents the introduction of grammatical errors.

\subsubsection{Fluency Consistency Constraint}

To evaluate linguistic fluency, we measure the change in perplexity (PPL)~\cite{radford2019language} between the original and poisoned sentences using a language model:

\begin{equation}
\Delta PPL = PPL(x') - PPL(x)
\label{eq:ppl_constraint}
\end{equation}

\subsubsection{Constraint Thresholds}

The thresholds for the linguistic constraints---semantic similarity ($\tau$), grammatical correctness ($\epsilon$), and fluency ($\Delta PPL$)---are summarized in Table~\ref{tab:constraint_thresholds}. These constraints are applied during trigger insertion to ensure that poisoned samples remain semantically consistent, grammatically valid, and fluent. The threshold values are selected based on validation experiments to achieve a balance between attack effectiveness and linguistic quality. Specifically, the semantic similarity threshold is set to $\tau = 0.92$ to ensure high semantic preservation between the original and poisoned sentences. The grammar error tolerance is limited to $\epsilon = 1$, restricting the introduction of grammatical inconsistencies. The fluency constraint enforces $\Delta PPL \leq 2.5$, ensuring that the generated sentences remain natural and readable.

Additionally, the maximum number of insertion attempts $A$ is set to 3 to balance effectiveness and computational efficiency. Increasing $A$ improves the likelihood of finding valid trigger positions but also increases computational cost, while lower values may fail to produce valid samples under the defined constraints.

\begin{table}[t]
\centering
\footnotesize
\renewcommand{\arraystretch}{1.25}
\caption{Threshold values used to enforce linguistic quality constraints during trigger insertion.}
\label{tab:constraint_thresholds}

\begin{tabular}{|l|c|c|}
\hline
\textbf{Constraint} & \textbf{Symbol} & \textbf{Value} \\
\hline
Semantic similarity threshold & $\tau$ & 0.92 \\
\hline
Grammar error tolerance & $\epsilon$ & 1.00 \\
\hline
Perplexity change monitoring & $\Delta PPL$ & $\leq 2.5$ \\
\hline
Maximum insertion attempts & $A$ & 3 \\
\hline
\end{tabular}
\end{table}

\subsection{Poisoning Attack Algorithm}

Algorithm~\ref{alg:combined_hin} presents the process of poisoning training samples generation. There are two modes that the algorithm can operate in, depending on $m$, where $m \in \{\texttt{token}, \texttt{sentence}\}$, i.e., whether to use tokens or sentence-level poisoning. Each sample from a set of candidates selected according to their non-target status will receive an injected trigger and be assessed using three language criteria: semantic similarity $\tau$, maximum tolerable grammatical errors $\epsilon$, and maximum tolerable perplexity deviation $\delta$. If the sample satisfies all three conditions, then it becomes part of the poisoned training set; otherwise, it remains intact. This may cause the actual poisoning rate to be lower than intended ($\lfloor p \cdot |S|\rfloor$), where $p$ stands for poison rate and $|S|$ denotes the size of the dataset. These criteria were experimentally validated.

Triggers generated in Algorithm~\ref{alg:combined_hin} are denoted as $t \sim P_{\text{token}}$ and $s \sim P_{\text{sent}}$.

\begin{algorithm}
\caption{Poisoning Attack Pipeline (HinToken / HinSent)}
\label{alg:combined_hin}
\begin{algorithmic}[1]

\REQUIRE Clean training set $S$, target label $y_t$, poison rate $p$,
attack mode $m \in \{\texttt{token}, \texttt{sentence}\}$,
trigger pools $P_{\text{token}}$/$P_{\text{sent}}$,
thresholds $\tau, \epsilon, \delta$, max attempts $A$

\ENSURE Poisoned training set $S_{\text{poisoned}}$,
success count $n_{\text{success}}$

\STATE Sample $\lfloor p \cdot |S| \rfloor$ indices from the set
$\{i \mid S[i].y \neq y_t\}$ and store in $C$

\STATE $S_{\text{poisoned}} \leftarrow S$;\;
$n_{\text{success}} \leftarrow 0$

\FOR{each $c \in C$}

    \STATE $x \leftarrow S[c].\text{text}$

    \FOR{$a = 1$ to $A$}

        \IF{$m = \texttt{token}$}

            \STATE $x' \leftarrow \text{Insert}(x,\;
            t \sim P_{\text{token}},\;
            j \sim \text{Uniform position in } x)$

        \ELSE

            \STATE $x' \leftarrow \text{Concat}(x,\;
            s \sim P_{\text{sent}})$

        \ENDIF

        \IF{$\text{USE}_{\text{sim}}(x,x') \geq \tau$
            \textbf{and} $\Delta\text{GE} \leq \epsilon$
            \textbf{and} $\Delta\text{PPL} \leq \delta$}

            \STATE $S_{\text{poisoned}}[c] \leftarrow (x', y_t)$;\;
            $n_{\text{success}} \mathrel{+}= 1$;\;
            \textbf{break}

        \ENDIF

    \ENDFOR

\ENDFOR

\STATE \RETURN $S_{\text{poisoned}},\; n_{\text{success}}$

\end{algorithmic}
\end{algorithm}

\subsection{Defense Algorithm}

Perplexity of the sentence $x$ from the suspicious sentences set is measured using large language models (LLMs). The set of models is defined as

\[
\mathcal{E} =
\{
\text{GPT-2},
\text{Gemma-2B},
\text{LLaMA-2}
\}.
\]

For each language model $m \in \mathcal{E}$, the corresponding perplexity score is calculated for the sentence $x$. Each model $m \in \mathcal{E}$ is viewed as an independent detector, whose effectiveness is estimated separately.

\begin{equation}
\label{eq:is_poisoned}
\text{is\_poisoned}_m(x) =
\begin{cases}
\text{True}, & \text{if } \text{PPL}_m(x) > \theta_m \\
\text{False}, & \text{otherwise}
\end{cases}
\end{equation}

Here $\text{PPL}_m(x)$ is the perplexity score of the sentence $x$ generated using the model $m$. $\theta_m$ is the threshold value which is calculated based on the distribution of perplexity scores for the clean validation set $D_{\text{clean}}$ by applying percentiles. Namely, $\theta_m$ is the $\alpha$-percentile of the perplexity distribution on the validation set $D_{\text{clean}}$:

\[
\theta_m =
\text{Percentile}_{\alpha}
\big(
\{
\text{PPL}_m(x) : x \in D_{\text{clean}}
\}
\big)
\]

We set $\alpha$ to $95$ which allows capturing only the upper end of the clean perplexity distribution. Clean sentences usually show consistent perplexity scores while poisoned or adversarial examples have high perplexity scores because of unnatural changes. Hence, sentences having a perplexity value above this threshold will be considered suspicious. It is worth mentioning that percentile thresholding is often utilized in out-of-distribution and anomaly detection scenarios. Finally, once the suspicious sentences are determined, we exclude them from our dataset.

\begin{algorithm}
\caption{Defense Employing Multiple Large Language Models}
\label{alg:perplexity_defender}
\begin{algorithmic}[1]

\REQUIRE Suspect dataset $D_{\text{suspect}}$,
clean validation set $D_{\text{clean}}$,
language model $LM$,
percentile threshold $\alpha$ (default: 95)

\ENSURE Filtered dataset $D_{\text{filtered}}$,
set of flagged samples $O$,
detection count $n_{\text{flagged}}$

\STATE \textit{// Phase 1: Compute perplexity}

\FOR{each sample $x_i \in D_{\text{suspect}}$}

    \STATE $p(x_i) \leftarrow \text{PPL}_{LM}(x_i)$

\ENDFOR

\STATE \textit{// Phase 2: Compute threshold from clean distribution}

\STATE $\theta \leftarrow \text{Percentile}_\alpha
\big(\{p(x_i) : x_i \in D_{\text{clean}}\}\big)$

\STATE \textit{// Phase 3: Flag and filter suspicious samples}

\STATE $O \leftarrow \emptyset$

\FOR{each sample $x_i \in D_{\text{suspect}}$}

    \IF{$p(x_i) > \theta$}

        \STATE $O \leftarrow O \cup \{i\}$

    \ENDIF

\ENDFOR

\STATE $D_{\text{filtered}} \leftarrow
D_{\text{suspect}} \setminus \{x_i : i \in O\}$

\STATE $n_{\text{flagged}} \leftarrow |O|$

\STATE \RETURN $D_{\text{filtered}},\; O,\; n_{\text{flagged}}$

\end{algorithmic}
\end{algorithm}

\subsection{Pipeline Design and Implementation}

\begin{figure*}[t]
    \centering
    \includegraphics[width=\textwidth]{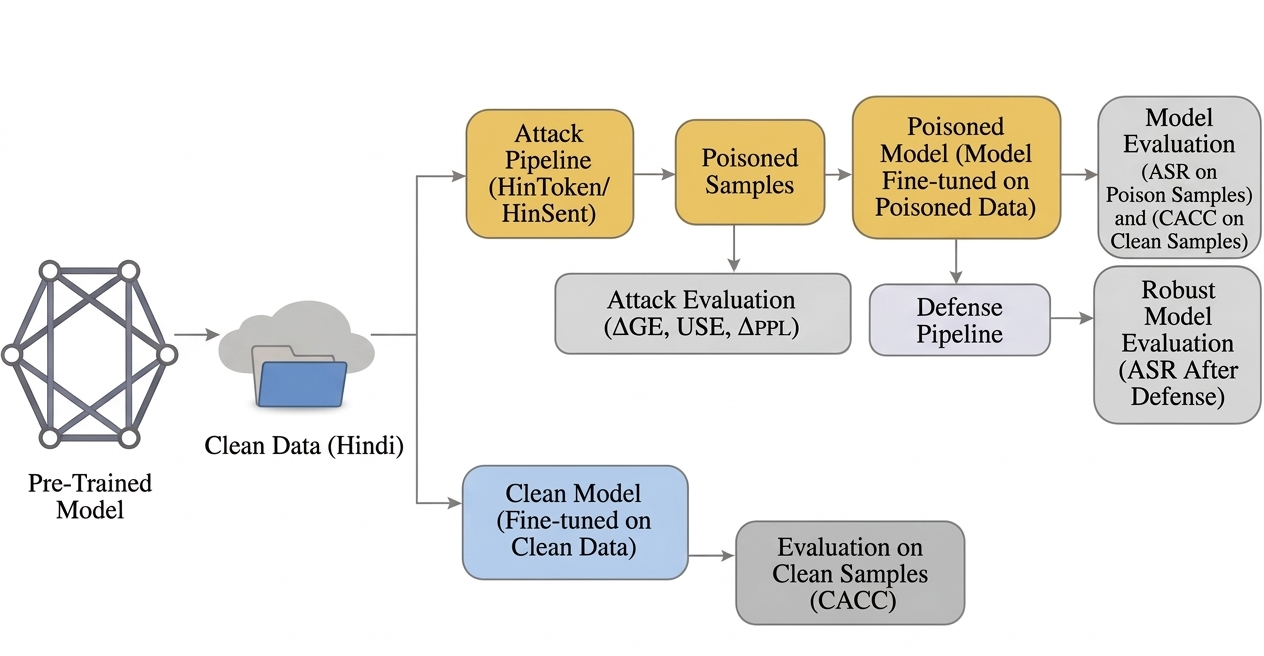}
    \caption{Attack--defense pipeline for Hindi models illustrating poisoning via HinToken and HinSent, evaluation using attack success rate (ASR) and clean accuracy (CACC), and improved robustness after applying defense mechanisms.}
    \label{fig:pipeline}
\end{figure*}

The proposed framework is based on a unified, modular pipeline that incorporates the processes of attack generation, model training, defense, and evaluation, as shown in Figure~\ref{fig:pipeline}. The pipeline begins with a pre-trained Transformer model (encoder base) and clean data in Hindi, after which it proceeds through a series of connected steps to enable the systematic exploration of vulnerabilities. From here, the pipeline diverges into two main data training configurations: a poisoned configuration and a clean baseline configuration.

\begin{itemize}

    \item \textbf{Poisoned Training Stage:}
    A portion of the training set is poisoned using HinToken/HinSent triggers to create poisoned samples with reverse target labels. Fine-tuning of the model on the mixed dataset results in a poisoned model that integrates the backdoor mechanism.

    \item \textbf{Clean Baseline Stage:}
    The pre-trained model is fine-tuned on clean data to create a clean baseline model. The clean model's performance on clean test samples is evaluated to measure the Clean Accuracy (CACC).

    \item \textbf{Vulnerability Evaluation:}
    The poisoned model's performance is evaluated using the Attack Success Rate (ASR) and CACC as well. In addition, linguistic stealth features such as $\Delta$GE, USE similarity, and $\Delta$PPL are calculated to measure the detectability of poisoned samples.

    \item \textbf{Defense and Robustness Evaluation:}
    An LLM-based defense approach is applied to filter suspicious samples from the training set. The model is fine-tuned on the filtered dataset, and robustness is evaluated by measuring the ASR after defense.

\end{itemize}

\noindent The modular design enables controlled comparison between clean, poisoned, and defended models under consistent preprocessing and training settings, facilitating comprehensive security evaluation.


\section{Experimental Setup}\label{sec4}

\subsection{Datasets}

We evaluate our approach on three datasets: IIT Patna Movie Reviews (Hindi)\footnote{\url{https://www.kaggle.com/datasets/warcoder/iit-patna-movie-reviews-hindi}}, Multilingual Toxicity Dataset\footnote{\url{https://huggingface.co/datasets/textdetox/multilingual_toxicity_dataset}}, and Hindi Sentiment Dataset\footnote{\url{https://www.kaggle.com/datasets/praths71018/hindi-sentiment-dataset}}. The IIT Patna Movie Reviews dataset is a Hindi movie review dataset annotated for sentiment classification with three classes. The Multilingual Toxicity Dataset contains text labeled for toxic and non-toxic content, representing a binary classification task. The Hindi Sentiment Dataset consists of sentences categorized into multiple sentiment classes, forming a multi-class classification setting. These datasets collectively cover both binary and multi-class classification tasks, enabling a comprehensive evaluation of the robustness of poisoning attacks. A summary of dataset statistics is presented in Table~\ref{tab:dataset-info}.

\begin{table*}[t]
\centering
\footnotesize
\renewcommand{\arraystretch}{1.25}
\caption{Summary of datasets used for experimentation.}
\label{tab:dataset-info}

\begin{tabular}{|l|c|c|c|c|}
\hline
\textbf{Dataset} & \textbf{Classes} & \textbf{Train} & \textbf{Valid} & \textbf{Test} \\
\hline
IIT Patna Movie Reviews (Hindi) & 3 & 2479 & 309 & 309 \\
\hline
Multilingual Toxicity Dataset & 2 & 1750 & 375 & 375 \\
\hline
Hindi Sentiment Dataset & 7 & 5600 & 1200 & 1200 \\
\hline
\end{tabular}
\end{table*}

\subsection{Victim Models}

We experimented with three transformer models commonly used for Hindi NLP: XLM-RoBERTa\footnote{\url{https://huggingface.co/FacebookAI/xlm-roberta-base}}, MuRIL\footnote{\url{https://huggingface.co/google/muril-base-cased}}, and Indic-BERT\footnote{\url{https://huggingface.co/ai4bharat/indic-bert}}. Each model differs in pretraining data and tokenization, providing diverse perspectives on attack vulnerability. Our pipeline supports any encoder-based transformer. Models were fine-tuned for their respective classification tasks.

\subsection{Training Details}

All victim models were pre-trained using publicly available pre-trained Transformer checkpoints and were fine-tuned for specific downstream Hindi text classification tasks. In order to make a fair comparison between various models, we used a set of hyperparameters that are standard for transformer model fine-tuning. This is summarized in Table~\ref{tab:training_details}. Poisoning is only applied to the training set.

\begin{table}[t]
\centering
\footnotesize
\renewcommand{\arraystretch}{1.2}
\caption{Hyperparameter configuration used for fine-tuning the victim models across all experiments.}
\label{tab:training_details}

\begin{tabular}{|l|c|}
\hline
\textbf{Hyperparameter} & \textbf{Value} \\
\hline
Optimizer & AdamW \\
\hline
Learning Rate & $2 \times 10^{-5}$ \\
\hline
Batch Size & 16 \\
\hline
Epochs & 15 \\
\hline
Warmup Ratio & 0.1 \\
\hline
Weight Decay & 0.01 \\
\hline
Max Sequence Length & 128 \\
\hline
Gradient Clipping & 1.0 \\
\hline
Poison Rate & 10\% \\
\hline
Random Seed & 42 \\
\hline
\end{tabular}
\end{table}

The hyperparameters are chosen following standard practices for fine-tuning transformer-based models to ensure stable convergence. We use the AdamW optimizer with a learning rate of $2 \times 10^{-5}$ and a batch size of 16. Regularization techniques, including weight decay and gradient clipping, are applied to mitigate overfitting. The maximum sequence length is set to 128, which is sufficient for sentence-level Hindi classification tasks. A unified hyperparameter configuration is adopted across all models to ensure fair comparison. Nevertheless, minor performance variations may arise due to differences in model architectures and tokenization schemes. Among the hyperparameters, the number of training epochs and the poisoning rate have the most significant impact on performance. Based on validation experiments, training for 15 epochs ensures convergence without overfitting, while a poisoning rate of 10\% provides a balance between attack effectiveness and clean model performance.

\subsection{Evaluation Metrics}
\label{sec:evaluation_metrics}

\begin{itemize}

\item \textbf{Clean Accuracy (CACC):}
Clean Accuracy measures the classification accuracy of the poisoned model on the original, unaltered test set. A high CACC value indicates that the poisoned model retains normal performance on clean inputs, which is essential for maintaining stealth under standard evaluation.

\item \textbf{Attack Success Rate (ASR):}
ASR quantifies the proportion of triggered test samples (drawn from non-target classes) that the poisoned model misclassifies as the attacker's desired target label~\cite{gu2019badnets, qi2021hidden}. A higher ASR reflects greater effectiveness of the backdoor mechanism.

\end{itemize}


\section{Results and Discussion}\label{sec5}

\subsection{Stealth and Linguistic Fidelity of Attacks}

This section discusses the stealth characteristics of the proposed poisoning attacks before moving on to their impact on the victim models. Table~\ref{tab:stealth_metrics_combined} presents a summary of the primary stealth metrics, change in perplexity ($\Delta$PPL), Universal Sentence Encoder similarity (USE), and change in grammatical errors ($\Delta$GE), for HinToken and HinSent, respectively.

\begin{table*}[t]
\centering
\footnotesize
\caption{Stealthiness and linguistic fidelity comparison between HinToken and HinSent attacks using different datasets. HinToken shows lower perturbation with higher semantic preservation than HinSent, which means HinToken is more effective with less perceptibility.}
\label{tab:stealth_metrics_combined}

\begin{tabular}{|l|l|c|c|c|}
\hline
& & \multicolumn{3}{c|}{\textbf{Stealth Metrics}} \\
\cline{3-5}
\textbf{Dataset} & \textbf{Attack} & $\boldsymbol{\Delta}\text{PPL}$ & \textbf{USE} & $\boldsymbol{\Delta}\text{GE}$ \\
\hline

Sentiment Analysis
& \multirow{3}{*}{HinToken}
& 0.0845
& 0.9973
& 0.91 \\
\cline{1-1}\cline{3-5}

Toxicity Dataset
&
& 1.5861
& 0.9909
& 0.89 \\
\cline{1-1}\cline{3-5}

IIT Patna Movie Reviews (Hindi)
&
& 0.0307
& 0.9965
& 0.90 \\
\hline

\textbf{Average}
&
& \textbf{0.5671}
& \textbf{0.9949}
& \textbf{0.90} \\
\hline

Sentiment Analysis
& \multirow{3}{*}{HinSent}
& 2.1011
& 0.9941
& 0.90 \\
\cline{1-1}\cline{3-5}

Toxicity Dataset
&
& 1.0281
& 0.9568
& 0.93 \\
\cline{1-1}\cline{3-5}

IIT Patna Movie Reviews (Hindi)
&
& 0.3575
& 0.9704
& 0.91 \\
\hline

\textbf{Average}
&
& \textbf{1.1622}
& \textbf{0.9738}
& \textbf{0.9133} \\
\hline

\end{tabular}
\end{table*}

The average values are reported to provide a dataset-agnostic summary of the attack behavior and to facilitate a direct comparison between HinToken and HinSent across different tasks. Since all datasets involve sentence-level classification and share the same evaluation metrics, averaging offers a meaningful view of overall trends.

\begin{figure*}[t]
\centering
\includegraphics[width=\textwidth]{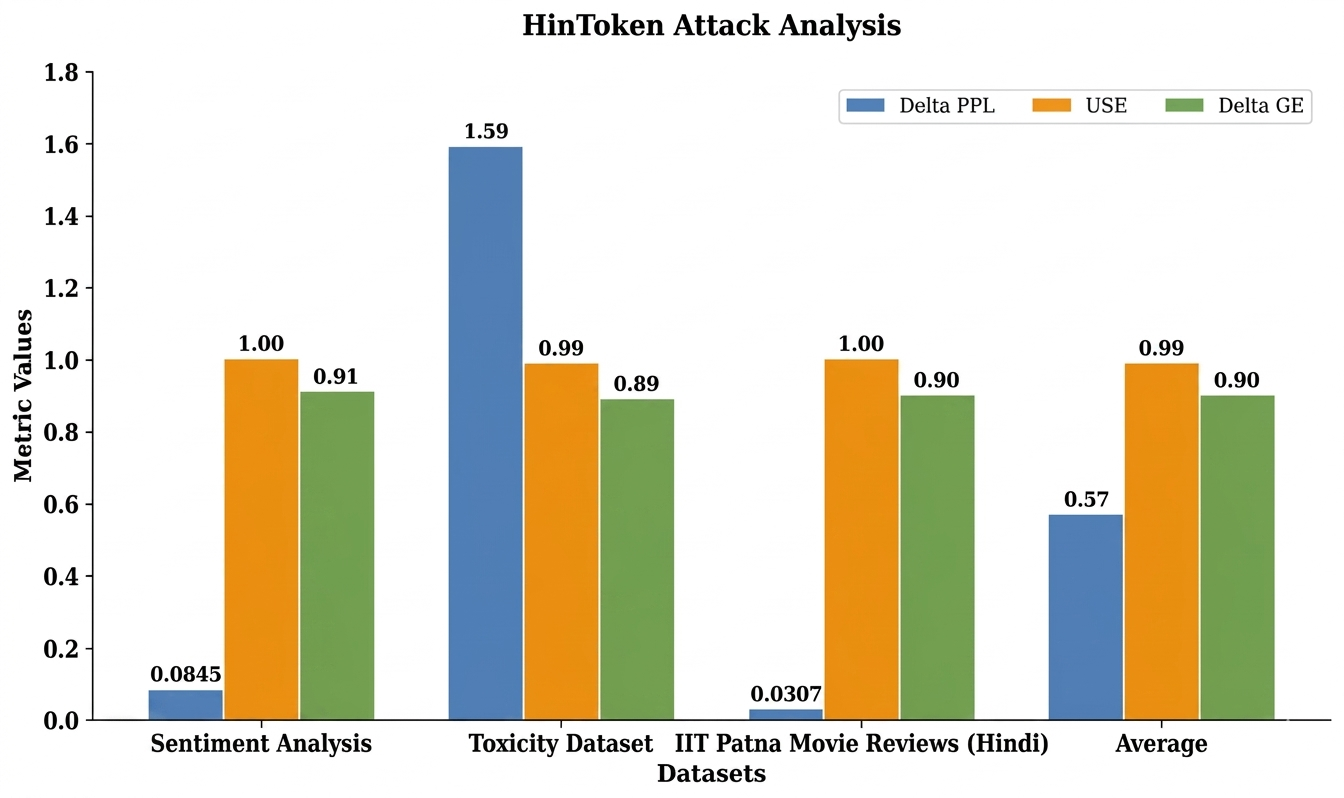}
\caption{Stealth characteristics of HinToken attacks across datasets, highlighting consistent fluency and semantic preservation under token-level perturbations.}
\label{fig:hintoken_metrics}
\end{figure*}

\begin{figure*}[t]
\centering
\includegraphics[width=\textwidth]{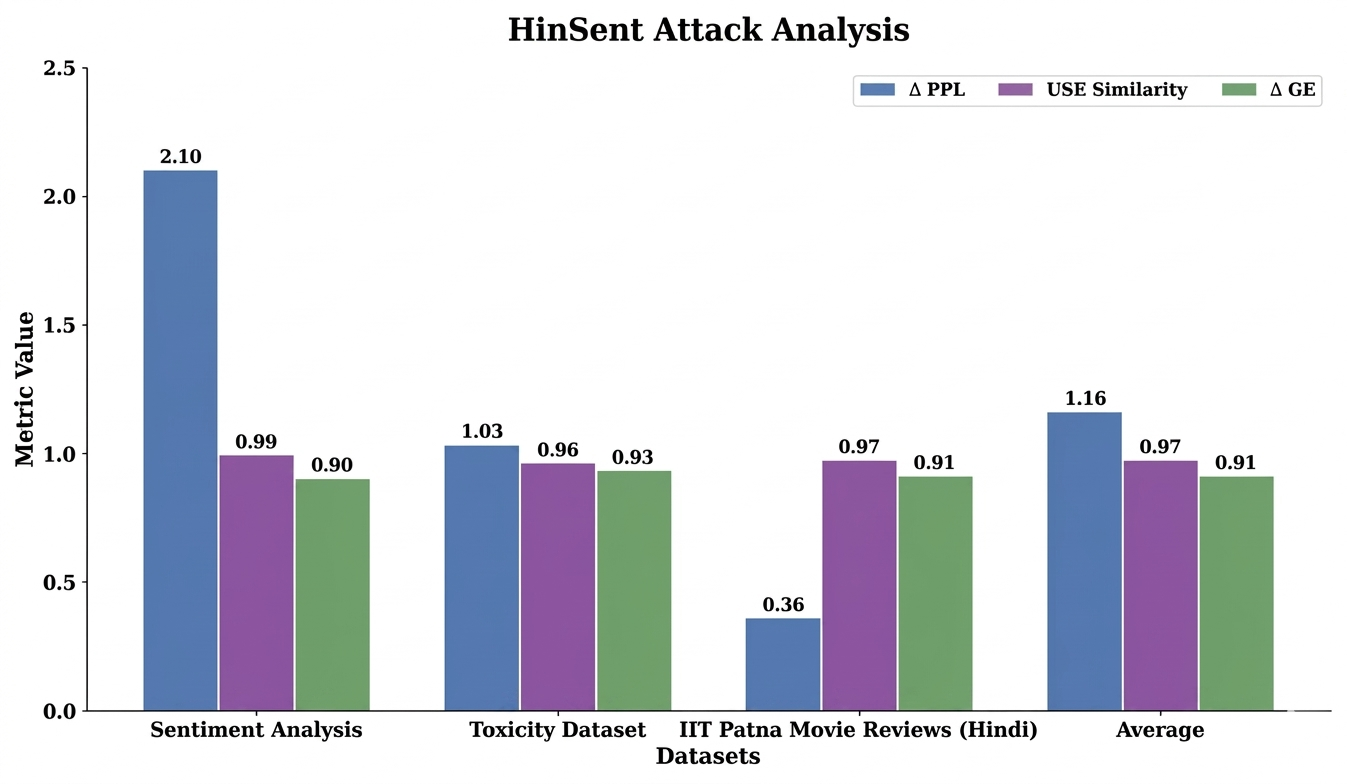}
\caption{Visualization of stealth metrics for HinSent attacks across datasets, showing change in perplexity ($\Delta$PPL), semantic similarity (USE), and grammatical error variation ($\Delta$GE). Compared to HinToken, higher $\Delta$PPL and slightly lower USE scores indicate that sentence-level perturbations introduce greater linguistic variation while still maintaining reasonable semantic consistency.}
\label{fig:hinsent_metrics}
\end{figure*}

The results show that HinToken attacks have a higher degree of stealth in preserving fluency and grammaticality compared to HinSent attacks. Notably, the lower $\Delta$PPL for HinToken implies that the poisoned inputs often have a level of fluency comparable to, or even higher than, their clean counterparts.

The results indicate that HinToken achieves higher stealth compared to HinSent, as reflected by consistently lower $\Delta$PPL and higher semantic similarity scores across datasets. In particular, the low $\Delta$PPL values suggest that HinToken preserves fluency close to that of clean inputs. While Table~\ref{tab:stealth_metrics_combined} provides exact quantitative values, Figures~\ref{fig:hintoken_metrics} and~\ref{fig:hinsent_metrics} illustrate the relative trends across datasets. The visualizations highlight the stability of HinToken in maintaining linguistic fidelity, whereas HinSent introduces comparatively larger perturbations, especially in terms of perplexity.

\subsection{Attack Effectiveness on Victim Models}

Prior to the execution of these defense mechanisms, the potency of these poisoning attacks was evaluated on the victim models using ASR and CACC metrics, as presented in Table~\ref{tab:attack_metrics_combined} for HinToken and HinSent triggers, respectively. Higher ASR values indicate successful activation of the poisoning attack, while constant CACC values indicate that model performance on clean data is unaffected. Among these models, XLM-RoBERTa and Indic-BERT models demonstrated high ASR values, indicating a higher likelihood of these models being vulnerable to both HinToken and HinSent poisoning attacks. On the other hand, MuRIL models demonstrated comparatively lower ASR values, indicating higher resilience to these poisoning attacks. This could be attributed to the unique design and pre-training data used by MuRIL models. Most importantly, constant CACC values for all models and datasets confirm that these poisoning attacks are effective and do not impact model performance on clean data.

\begin{table*}[t]
\centering
\footnotesize
\caption{Attack effectiveness of HinToken and HinSent triggers across victim models and datasets measured using ASR and CACC. The results show that both attacks achieve high ASR across models, with XLM-RoBERTa and Indic-BERT being more vulnerable, while MuRIL exhibits relatively higher robustness.}
\label{tab:attack_metrics_combined}

\begin{tabular}{|l|l|c|c|c|c|}
\hline
& & \multicolumn{2}{c|}{\textbf{HinToken}} & \multicolumn{2}{c|}{\textbf{HinSent}} \\
\cline{3-6}
\textbf{Dataset} & \textbf{Victim Model} & \textbf{ASR} & \textbf{CACC} & \textbf{ASR} & \textbf{CACC} \\
\hline

Sentiment Analysis
& \multirow{3}{*}{XLM-RoBERTa}
& 0.6640 & 0.9160 & 0.4484 & 0.9192 \\
\cline{1-1}\cline{3-6}

Toxicity Dataset
&
& 0.3746 & 0.9850 & 0.4121 & 0.9850 \\
\cline{1-1}\cline{3-6}

IIT Patna Movie Reviews (Hindi)
&
& 0.6840 & 0.8980 & 0.8172 & 0.9935 \\
\hline

Sentiment Analysis
& \multirow{3}{*}{MuRIL}
& 0.4484 & 0.9192 & 0.9361 & 0.8915 \\
\cline{1-1}\cline{3-6}

Toxicity Dataset
&
& 0.4282 & 0.9533 & 0.2920 & 0.9120 \\
\cline{1-1}\cline{3-6}

IIT Patna Movie Reviews (Hindi)
&
& 0.6400 & 0.9060 & 0.6400 & 0.9060 \\
\hline

Sentiment Analysis
& \multirow{3}{*}{Indic-BERT}
& 0.4433 & 0.5417 & 0.9894 & 0.9161 \\
\cline{1-1}\cline{3-6}

Toxicity Dataset
&
& 0.4141 & 0.8442 & 0.4000 & 0.9400 \\
\cline{1-1}\cline{3-6}

IIT Patna Movie Reviews (Hindi)
&
& 0.6840 & 0.8980 & 0.6700 & 0.8700 \\
\hline

\end{tabular}
\end{table*}

\subsection{Comparative Analysis of Attack Success Across Models and Datasets}

\begin{figure*}[t]
\centering
\includegraphics[width=\textwidth]{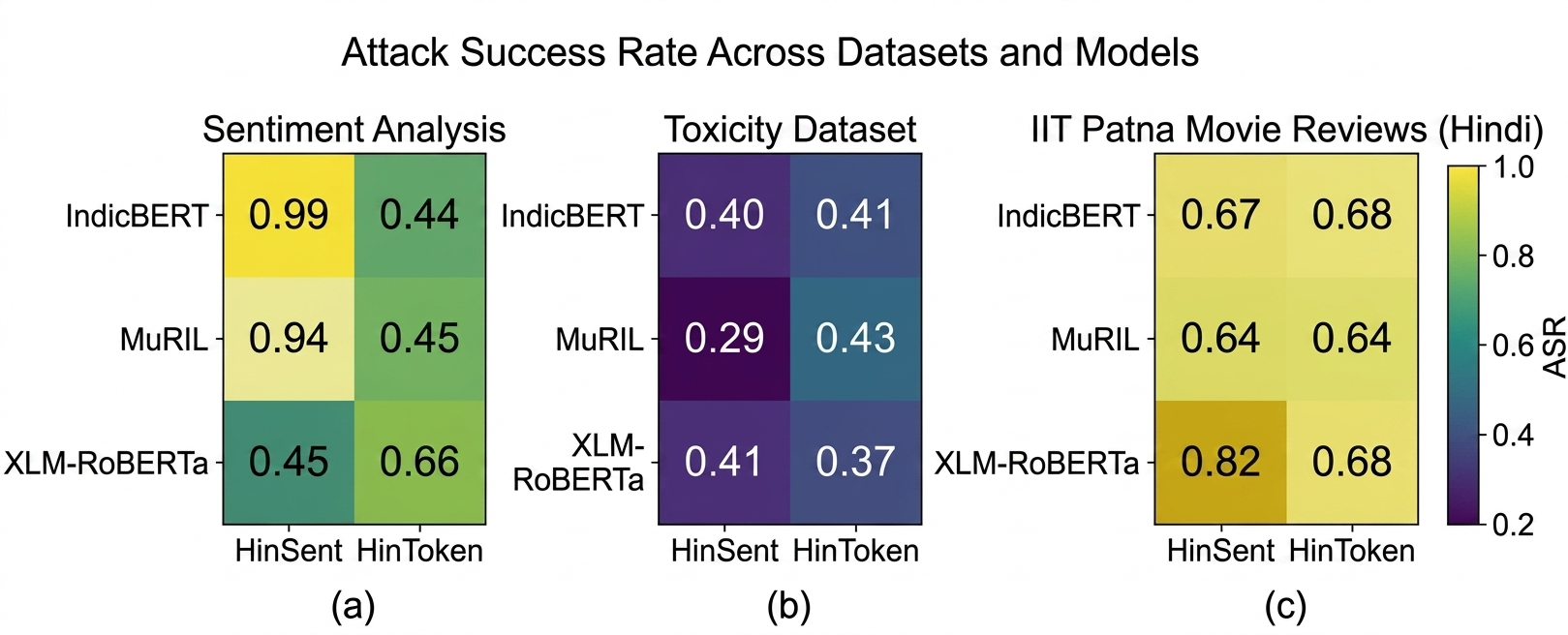}
\caption{Heatmap representation of Attack Success Rate (ASR) across victim models and datasets for HinSent and HinToken attacks, demonstrating dataset-specific robustness and differences in trigger effectiveness across model architectures.}
\label{fig:asr_heatmap}
\end{figure*}

To better understand the patterns of cross-dataset vulnerability, we also visualize the attack success rates (ASR) of the HinSent and HinToken triggers across the victim models in Figure~\ref{fig:asr_heatmap}. It can be seen that while HinSent generally has higher ASR for sentiment-focused datasets in most cases, HinToken presents a relatively more consistent pattern over datasets. It is also observed that XLM-RoBERTa shows higher ASR over the IIT Patna movie review dataset, while all models have lower attack success rates over the toxicity dataset, making them more resistant to poisoning triggers for the dataset.

\subsection{Post-Defense Attack Success Rate}

Table~\ref{tab:hintoken_defense_combined} and Table~\ref{tab:hinsent_defense_combined} show the Attack Success Rate (ASR) before and after applying the defense on all victim models, datasets, and defense language models (GPT-2, Gemma-2B, and LLaMA-2) for both HinToken and HinSent attacks. From all experimental settings, it is evident that the defense mechanism effectively decreases ASR. Across the evaluated defense models, the average ASR reduction ranges from approximately 33\% to 58\%, depending on the attack type and defense model. This proves that the defense is strong regardless of trigger construction and victim model architectures.

In Table~\ref{tab:hintoken_defense_combined} for HinToken attacks, it is evident that the defense significantly decreases ASR for all datasets and models. On average, ASR is reduced from 0.5311 to 0.228 when using Gemma-2B, outperforming GPT-2 and LLaMA-2.

In Table~\ref{tab:hinsent_defense_combined} for HinSent attacks, it is evident that the defense significantly decreases ASR for all datasets and models. In this case, LLaMA-2 outperforms GPT-2 and Gemma-2B by a significant margin, reducing ASR from 0.6228 to 0.263, compared to 0.416 for GPT-2 and 0.365 for Gemma-2B. These improvements are more significant when ASR is high, proving that the defense is effective in dealing with sentence-level triggers.

\begin{table*}[t]
\centering
\footnotesize
\caption{ASR before and after applying the proposed defense for HinToken attacks across all victim models and datasets. The results show a consistent reduction in ASR across all configurations, with Gemma-2B achieving the strongest suppression compared to GPT-2 and LLaMA-2.}
\label{tab:hintoken_defense_combined}

\begin{tabular}{|l|l|c|c|c|c|}
\hline
& & \textbf{ASR After} & \multicolumn{3}{c|}{\textbf{ASR After Defense}} \\
\cline{4-6}
\textbf{Dataset} & \textbf{Victim Model} & \textbf{Attack}
& \textbf{GPT-2} & \textbf{Gemma-2B} & \textbf{LLaMA-2} \\
\hline

Sentiment Analysis
& \multirow{3}{*}{XLM-RoBERTa}
& 0.6640 & 0.452 & 0.400 & 0.415 \\
\cline{1-1}\cline{3-6}

Toxicity Dataset
&
& 0.3746 & 0.155 & 0.100 & 0.235 \\
\cline{1-1}\cline{3-6}

IIT Patna Movie Reviews (Hindi)
&
& 0.6840 & 0.342 & 0.220 & 0.430 \\
\hline

Sentiment Analysis
& \multirow{3}{*}{MuRIL}
& 0.4484 & 0.205 & 0.180 & 0.295 \\
\cline{1-1}\cline{3-6}

Toxicity Dataset
&
& 0.4282 & 0.291 & 0.170 & 0.275 \\
\cline{1-1}\cline{3-6}

IIT Patna Movie Reviews (Hindi)
&
& 0.6400 & 0.335 & 0.310 & 0.325 \\
\hline

Sentiment Analysis
& \multirow{3}{*}{Indic-BERT}
& 0.4433 & 0.302 & 0.190 & 0.310 \\
\cline{1-1}\cline{3-6}

Toxicity Dataset
&
& 0.4141 & 0.282 & 0.260 & 0.185 \\
\cline{1-1}\cline{3-6}

IIT Patna Movie Reviews (Hindi)
&
& 0.6840 & 0.465 & 0.230 & 0.340 \\
\hline

\textbf{Average}
&
& \textbf{0.5311}
& \textbf{0.314}
& \textbf{0.228}
& \textbf{0.312} \\
\hline

\end{tabular}
\end{table*}

\begin{table*}[t]
\centering
\footnotesize
\caption{ASR before and after applying the proposed defense for HinSent attacks across all victim models and datasets. The defense significantly reduces ASR across all settings, with larger reductions observed in high-ASR scenarios. Among the evaluated defense models, LLaMA-2 achieves the lowest average ASR.}
\label{tab:hinsent_defense_combined}

\begin{tabular}{|l|l|c|c|c|c|}
\hline
& & \textbf{ASR After} & \multicolumn{3}{c|}{\textbf{ASR After Defense}} \\
\cline{4-6}
\textbf{Dataset} & \textbf{Victim Model} & \textbf{Attack}
& \textbf{GPT-2} & \textbf{Gemma-2B} & \textbf{LLaMA-2} \\
\hline

Sentiment Analysis
& \multirow{3}{*}{XLM-RoBERTa}
& 0.4484 & 0.305 & 0.210 & 0.320 \\
\cline{1-1}\cline{3-6}

Toxicity Dataset
&
& 0.4121 & 0.270 & 0.250 & 0.261 \\
\cline{1-1}\cline{3-6}

IIT Patna Movie Reviews (Hindi)
&
& 0.8172 & 0.680 & 0.580 & 0.200 \\
\hline

Sentiment Analysis
& \multirow{3}{*}{MuRIL}
& 0.9361 & 0.580 & 0.500 & 0.310 \\
\cline{1-1}\cline{3-6}

Toxicity Dataset
&
& 0.2920 & 0.198 & 0.150 & 0.125 \\
\cline{1-1}\cline{3-6}

IIT Patna Movie Reviews (Hindi)
&
& 0.6400 & 0.435 & 0.400 & 0.215 \\
\hline

Sentiment Analysis
& \multirow{3}{*}{Indic-BERT}
& 0.9894 & 0.572 & 0.560 & 0.375 \\
\cline{1-1}\cline{3-6}

Toxicity Dataset
&
& 0.4000 & 0.250 & 0.220 & 0.135 \\
\cline{1-1}\cline{3-6}

IIT Patna Movie Reviews (Hindi)
&
& 0.6700 & 0.460 & 0.420 & 0.430 \\
\hline

\textbf{Average}
&
& \textbf{0.6228}
& \textbf{0.416}
& \textbf{0.365}
& \textbf{0.263} \\
\hline

\end{tabular}
\end{table*}

Overall, the results conclude that the efficacy of this defense mechanism depends on the models and types of attacks used in this context, where Gemma-2B is found to be more effective in comparison to token-level attacks, while LLaMA-2 is found to be more effective in comparison to sentence-level attacks.


\section{Conclusion}

This work investigated the susceptibility of Transformer-based text classifiers to data-poisoning attacks in the Hindi language setting, an area that has received comparatively little attention despite the scale of the Hindi-speaking user base. Two stealthy poisoning strategies were introduced: HinToken, which embeds a low-frequency Hindi word as a covert trigger, and HinSent, which appends a short, naturally phrased Hindi sentence to induce misclassification. Both approaches were designed to preserve the linguistic integrity of the input while still reliably steering model predictions toward an attacker-chosen label.

The stealth analysis showed that HinToken introduces only marginal disruption to fluency and semantic content, with an average USE similarity of 0.9949 and a change in perplexity as low as 0.0307 on the IIT Patna Movie Reviews dataset, making poisoned samples difficult to distinguish from clean ones. HinSent, while inherently a larger textual modification, still preserved a strong average semantic similarity of 0.9738, confirming that even sentence-level insertions can remain linguistically coherent.

In terms of attack potency, both strategies proved highly effective against the evaluated victim models. HinSent reached an attack success rate as high as 98.94\% on the Indic-BERT sentiment classifier, while HinToken achieved up to 68.40\% on the IIT Patna Movie Reviews dataset, in both cases without a corresponding drop in clean accuracy — indicating that a poisoned model can behave indistinguishably from a clean one under standard evaluation. Across models, Indic-BERT was the most consistently vulnerable to both attack types, while XLM-RoBERTa and MuRIL exhibited comparatively lower and more similar susceptibility to one another, suggesting that pretraining corpus composition and tokenization strategy play a meaningful role in a model's exposure to trigger-based poisoning. Task-wise, sentiment classification was generally more affected by HinSent-style triggers, whereas the toxicity classification task showed greater resistance, although this pattern varied somewhat by dataset.

To counter these vulnerabilities, an LLM-based filtering defense was developed that flags training samples with anomalously high perplexity relative to a clean reference distribution, requiring no prior knowledge of the trigger patterns used during poisoning. Using GPT-2, Gemma-2B, and LLaMA-2 as independent perplexity estimators, the defense reduced the average attack success rate for HinToken from 0.5311 to approximately 0.2847 across the three models — an average reduction of roughly 46.4\% — and for HinSent from 0.6228 to approximately 0.3480, a reduction of roughly 44.1\%. Gemma-2B was the strongest defender against HinToken, while LLaMA-2 was most effective against HinSent, pointing to complementary strengths across language models that a multi-model defense strategy can exploit. 
Overall, this study makes three main contributions. First, it introduces HinToken and HinSent, two linguistically informed poisoning attacks tailored to the morphological and syntactic properties of Hindi rather than adapted from generic, language-agnostic trigger designs. Second, it demonstrates that perplexity-based filtering using multiple large language models offers a practical, trigger-agnostic defense that generalizes across attack types without requiring labeled poisoned examples. Third, it establishes an end-to-end attack-defense-evaluation pipeline that can serve as a reusable benchmark for future security research on Hindi and other Indian-language NLP systems.

These findings underscore that robustness evaluation for NLP systems cannot be assumed to transfer uniformly from high-resource languages to low-resource, morphologically rich languages such as Hindi. Language-aware threat modeling, together with defenses that account for linguistic structure, will be necessary to build text classification systems that are both secure and trustworthy for the Indian language ecosystem.


\section*{Acknowledgments}
This work was carried out as part of the project funded by the National Supercomputing Mission (NSM), a flagship initiative supported by the Ministry of Electronics and Information Technology (MeitY), Government of India.


\bibliographystyle{IEEEtran}
\bibliography{references}

\end{document}